\documentclass[letterpaper, 10 pt, conference]{ieeeconf}  

\IEEEoverridecommandlockouts                              
\usepackage[colorinlistoftodos]{todonotes}
\usepackage{multirow}
\usepackage{graphicx}
\usepackage{amsfonts}
\usepackage{comment}
\usepackage{hyperref}
\usepackage{url}

\usepackage[ruled,vlined]{algorithm2e}
\usepackage{amsmath}
\usepackage{algpseudocode}
\usepackage{caption}
\usepackage{subcaption}
\usepackage{dblfloatfix}
\newlength{\twosubht}
\newsavebox{\twosubbox}

\usepackage{graphicx}
\usepackage{subcaption}
\usepackage{tikz}
\usepackage{multicol}
\usepackage{newtxtext}
\usepackage{newtxmath}
\usepackage{color,soul}

\usepackage{booktabs} 
\usepackage{xparse}   
\NewDocumentCommand{\rot}{O{45} O{1em} m}{\makebox[#2][l]{\rotatebox{#1}{#3}}}%
\title{\LARGE \bf
 6D Assembly Pose Estimation by Point Cloud Registration for Robot Manipulation  
}

\author{Kulunu Samarawickrama$^{1}$, Gaurang Sharma$^{1}$, Alexandre Angleraud$^{1}$ and Roel Pieters$^{1}$
\thanks{$^{1}$Automation Technology and Mechanical Engineering, Tampere University, 33720, Tampere, Finland, {\tt\small firstname.surname@tuni.fi}}%
}

\begin{document}

\maketitle
\thispagestyle{empty}
\pagestyle{empty}

\begin{abstract}

The demands on robotic manipulation skills to perform challenging tasks have drastically increased in recent times. To perform these tasks with dexterity, robots require perception tools to understand the scene and extract useful information that transforms to robot control inputs. To this end, recent research has introduced various object pose estimation and grasp pose detection methods that yield precise results. Assembly pose estimation is a secondary yet highly desirable skill in robotic assembling as it requires more detailed information on object placement as compared to bin picking and pick-and-place tasks. However, it has been often overlooked in research due to the complexity of integration in an agile framework. To address this issue, we propose an assembly pose estimation method with RGB-D input and 3D CAD models of the associated objects. The framework consists of semantic segmentation of the scene and registering point clouds of local surfaces against target point clouds derived from CAD models to estimate 6D poses. We show that our method can deliver sufficient accuracy for assembling object assemblies using evaluation metrics and demonstrations. The source code and dataset for the work can be found at: \textit{\url{https://github.com/KulunuOS/6DAPose}}


\end{abstract}

\section{Introduction}\label{sec:intro}
The ability to grasp an object successfully is a highly desired skill of robotic manipulators in applications such as industrial manufacturing, mining, space explorations, etc. The advancement of modern robot manipulators with multiple degrees of freedom, precise sensing and control capabilities highly encourages the development of the robot's cognition skills in order to execute dexterous tasks in above fields either autonomously or with human collaboration. Several research publications have demonstrated that vision-based deep learning  and reinforcement learning can yield successful results in endowing cognitive skills such as object grasping \cite{kleeberger2020survey,Liang}.
\begin{figure}[t]
  \centering
  \includegraphics[width=0.5\textwidth]{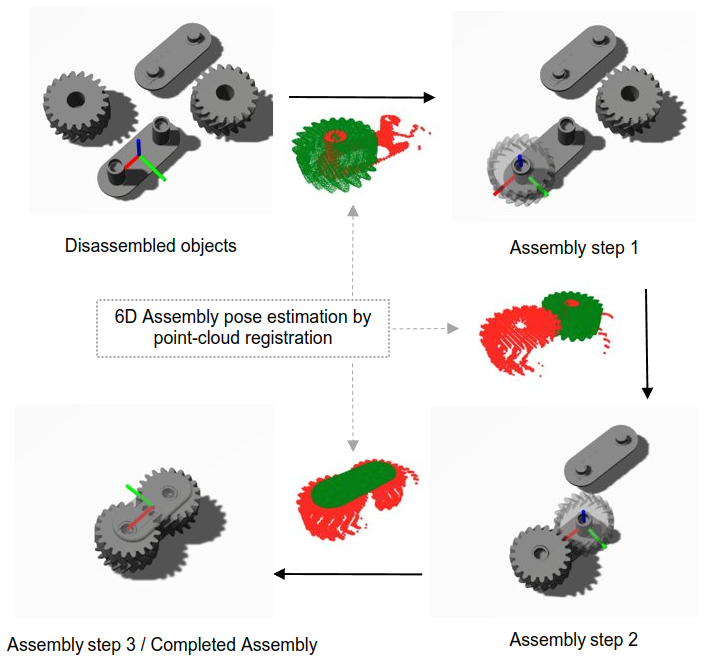}
  \caption{Overview of assembly pose estimation pipeline depicting for each step the base assembly point cloud in red and the estimated assembly point cloud (object pose) in green.\vspace{0mm} 
  }
    \label{fig:overview} 
\end{figure}
In that regard, the state of the art on robotic manipulation is heavily focused on object pose estimation and grasp pose detection \cite{mohammed2020review, Du2021}. Pose estimation via point cloud based deep learning methods such as PVN3D \cite{He_2021_CVPR} and FFB6D \cite{He_2020_CVPR} have shown great accuracy in 6D pose estimation with fast inference times, which are ideal characteristics for robotic grasping applications. However, object pose alone is not sufficient to perform a successful grasp. The robot end-effector should reach a pose that produces a fail-proof grasping of the object, which is a different definition compared to the object pose. Grasp pose for an object can be sampled either analytically or using human expertise. Some analytical methods \cite{Kleeberger,Xu2023,Fast-geometry-based} have proposed deriving grasp pose by analyzing object model and depth information acquired from a sensor. Although these methods can produce feasible grasp poses for a single object, they fail when objects are in clutter. GraspNet  \cite{mousavian2019graspnet} introduced a model-agnostic method to directly produce grasp candidates on any observed unknown partial point cloud. This method produces multiple grasp candidates and prunes unfeasible candidates in an end-to-end pipeline which makes it highly suitable for bin picking, sorting and inspection tasks.
High computational resource consumption of these methods are, however, an undesirable feature for robotic applications. 

However, there exists a disparity in research for precise robotic assembly, as the majority of the existing studies on robotic assembly focus only on initial object pose estimation. Robotic assembly requires the additional estimation of object placement pose (see Fig. \ref{fig:overview}). We use the synonymous term 'assembly pose' henceforth in this paper.  Assembly pose can be described as the pose of the end-effector that results in the successful placement of a grasped object that satisfies assembly constraints. The primary constraint that defines accuracy of assembly is the relative pose between objects of the assembly. Other assembly constraints vary depending on the nature of the task and the task environment, such as specific placement direction or motion, but is not considered in this work.

To address the issue of assembly pose estimation, we propose an approach that can be integrated into a robotic assembly framework without compromising the agility. As the final step of a robotic manipulation framework, an efficient method should well utilize the inferred information from object pose estimation and grasp pose detection steps, offer high accuracy, and not compromise the computational load and inference times. First, the proposed method includes a semantic segmentation module that estimates a per-pixel object label. Second, the segmentation results and RGB-D images are utilized to perform point cloud registration of a pair of source and target point clouds. As an object assembly may consist of several assembly steps, assembly poses are estimated as a sequential process, but separately for each individual assembly step. Pose estimation does therefore not rely on previous steps, besides the assumption of a correct previous assembly step, represented in the CAD model. 
The proposed method can be integrated with a framework consisting of existing pose estimation and grasp detection methods without additional model training. 

The contributions of this work are as follow:
\begin{enumerate}
    \item Adaptation of state of the art object pose estimation datasets for assembly pose estimation
    \item An effective method to generate source point clouds using CAD models
    \item An iterative pipeline to estimate assembly poses for object assemblies with multiple objects
    \item An evaluation of efficiency of point cloud registration as a method of assembly pose estimation
\end{enumerate}

The remainder of this paper is organized as follows. In Section \ref{sec:related_work} we provide an overview of related work. Section \ref{sec:data_gen} presents our proposed method, which is evaluated in Section \ref{sec:evaluation} and discussed in Section \ref{sec:discussion}. Section \ref{sec:conclusion} concludes the work.

\section{Related Work}\label{sec:related_work}
An early attempt at combining image segmentation and point cloud registration for pose estimation and robotic grasping featured in the Amazon bin picking challenge in 2016 \cite{amazon}. Zeng et al. takes an RGB-D input of the bin and outputs 6D object poses. The estimated object poses are converted to grasp poses to be grasped by a parallel gripper. They implemented a Fully connected Neural Network with VGG architecture for object segmentation in their method with a large dataset of around 130,000 images. Despite the fact that segmentation networks have vastly improved over time since 2016, their results show that the heavy occlusive and cluttered nature of bin picking tasks negatively affected point cloud registration, thereby increasing the pose estimation error. Besides, they constrain the source point cloud by orienting it towards the optical axis of the camera in the pose initializing step. The resulting ambiguity of rotation around the axis could result in an initial error that propagates through ICP refinement. Some later research proposed a solution to the problem by introducing model libraries containing samples of partial point clouds as seen from different view points for each unique object category.

SegICP \cite{SegICP} proposed another per-pixel instance segmentation module and an ICP registration method for the same purpose. They adopt Segnet \cite {Segnet}, a convolutional neural network, for the semantic segmentation step in their pipeline. In the point cloud registration step they align a source point cloud against a target point cloud retrieved from a model library. Ultimately, they perform an exhaustive search for the best pose estimation by comparing the ICP result against all samples in the model library for each object category.

Some further studies have been done using SegICP as a base concept. Xu et al. \cite{xu2019} proposed a pose estimation pipeline with a custom segmentation network (FCN-artous-2s) and modified ICP algorithm for a grasp manipulation task. The modified ICP algorithm addresses lack of one-to-one correspondence points in partial point clouds of the target and source. Wang et al. \cite{wang2019pose} proposed a similar pose estimation method in their pipeline for a robotic spray painting application. They implement an additional network to estimate the initial orientation, which facilitates retrieval of the most accurate point cloud sample from the model library.  However, the additional iterations and inference steps in these methods results in prolonged total inference times.

In contrast to above work, we perform real-time rendering of the target point cloud in the partial observable state without sampling views and generating model libraries. This eliminates naive assumptions in both pose initialization and exhaustive, repetitive ICP calculations. The evaluation metric used to measure registration accuracy directly affects the pose estimation accuracy. Zeng et al. \cite{amazon} do not evaluate the registration result and SegICP uses a unique point correspondence matching criteria for evaluation of the registration result. We use two different evaluation metrics, namely inlier RMSE (Root Mean Square error) and a fitness score to analyze registration accuracy and reflect its propagation towards pose estimation accuracy in our work.

\section{Proposed Method}\label{sec:data_gen}

\begin{figure*}[t!]
    \centering
        \centering
        \includegraphics[width=\textwidth]{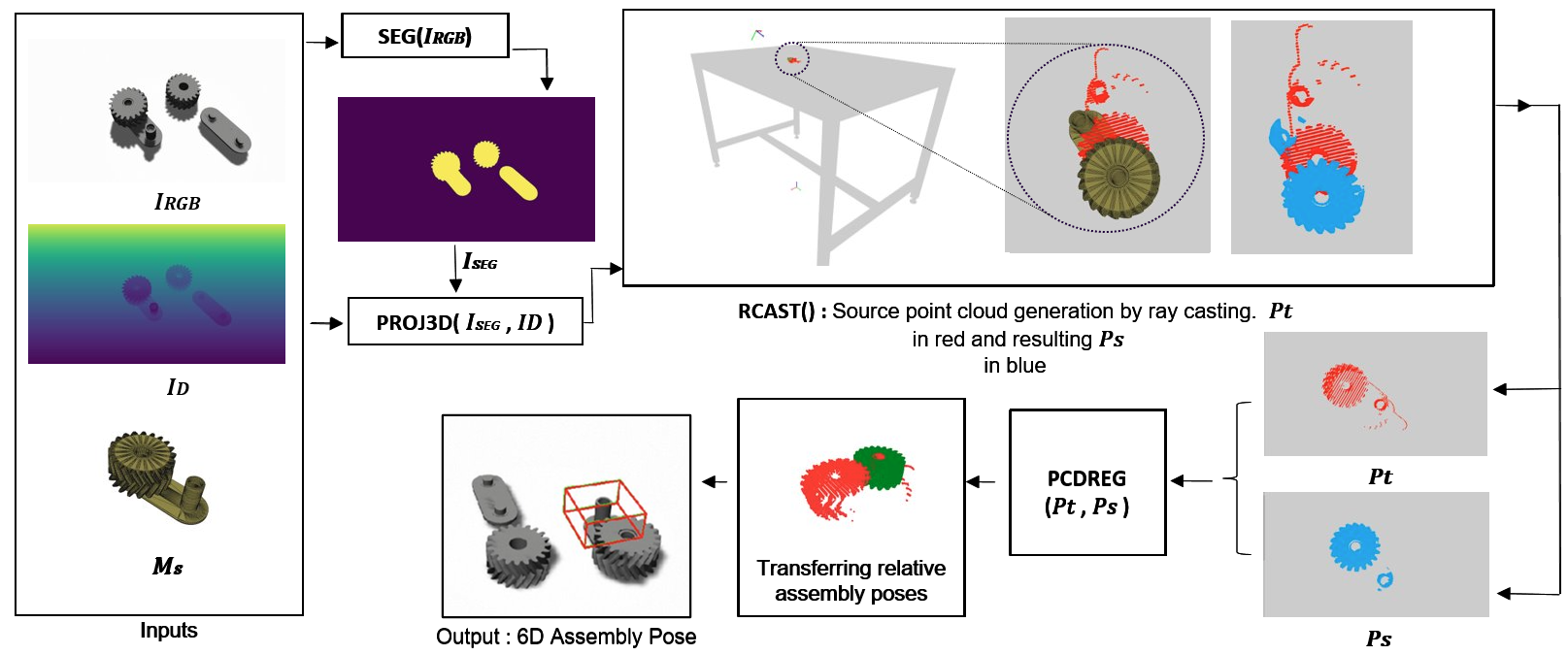}
    
    \caption{Proposed framework for 6D assembly pose estimation \label{fig:framework}\vspace{0mm}}
\end{figure*}

We approach the object assembling task as an iterative process of several sub-assembling steps as described in Fig. \ref{fig:overview}. For the proposed method we assume that the order of assembling and pose of each object with respect to the base object are known. The base object can be defined as the completed assembly of objects at each step as described in Fig. \ref{fig:framework}. More complex and larger object assemblies can be divided into several sub-assemblies and corresponding base objects. Therefore it can be assumed that all objects assemblies can be assembled in above described manner.

An assembly pose estimation procedure that can be integrated to a generic vision based robotic assembling framework is presented in Fig. \ref{fig:framework}. The first step of the framework is to understand the scene and locate the objects. The second step is to produce a fail-proof grasp for each object. These two tasks can be achieved by object pose estimation and grasp pose estimation algorithms. Our proposed method applies in the third step where a successfully grasped object is assembled to satisfy assembly constrains. The assembly pose estimation algorithm is described in Alg. \ref{alg:assembly_pose}. 

\begin{algorithm}[!th]
\SetAlgoLined
\SetKwInOut{KwFuncs}{Functions}
\SetKwInOut{KwInput}{Input}
\SetKwInOut{KwOutput}{Output}
\SetKwInOut{KwParams}{Parameters}
\KwParams{ }
\KwInput{$I_{RGB}, I_{D}$, $M_{s = 1,2,3 .. n}$}
\KwOutput {$T_w^a$}
\KwFuncs{ SEG() ; Instance Semantic Segmentation\\
          PROJ3D() ; 3D projection of $I_{D}$ \\
          TFORM() ; $SE(3)$ Transformation\\
          RCAST() ; Raycasting in simulation \\ 
          PCDREG() ; Pointcloud Registration \\ 
         }
\ForEach{ Assembly step}{
            $I_{SEG}$  $\leftarrow$ SEG($I_{RGB})$\\
            $P_{t}$ $\leftarrow$ PROJ3D( $I_{SEG}$, $I_{D}$)\\ 
            $P_{s}$ $\leftarrow$ RCAST(TFORM($M_{s}$,$P_{t}$))\\
            $T_b^w$ $\leftarrow$ PCDREG($P_{t}$ , $P_{s}$)\\
            $T_w^a$ =  $T_b^a$   $T_w^b$ \\
            
}

\caption{Assembly pose estimation}
\label{alg:assembly_pose}
\end{algorithm}

 
\subsection{Object and Grasp Pose Detection}
Object pose locates the object in robot coordinate frame. Grasp pose defines the pose of robot's end effector while producing a fail-proof grasp of the object. A successful grasp is a prerequisite for a successful assembly and depends on both above factors. However, object grasping is not the main focus of our work. The pipeline provides the opportunity to utilize a grasp pose detection method of preference. A module that first estimates the object pose and then derives grasp poses relative to object pose is more suitable for the task. Alternatively, point cloud registration result can be used to detect object pose as well when relative grasp pose information is available as in our demonstration. 


\subsection{Assembly Pose Estimation}  
For a robotic assembling task, assembly pose estimation can be interpreted as the estimation of rotation and translation of the assembly object frame with respect to the robot base frame. An assembly pose must satisfy certain assembly constraints to create a precise assembly which varies depending on the task. Often in industrial object assemblies, the assembly pose is constrained by the relative pose of assembly object with respect to a base object of an assembly or an assembly surface. 

In Alg. \ref{alg:assembly_pose}, each base object is represented by a source mesh $M_{s}$. A prior knowledge $T_b^a$ defines the relative pose from base object $b$ to assembly object $a$ by the homogeneous transformation matrix  $T_b^a$. The output of the algorithm is the homogeneous transformation matrix $T_w^a$ which estimates the relative assembly pose of assembly object  to world frame $w$.

\subsubsection*{Instance semantic segmentation}
The first function in algorithm $SEG()$, an instance semantic segmentation module trained on RGB images $I_{RGB}$ estimates per-pixel assembly object label for a given scene. A segmentation mask of the base object $I_{SEG}$ in the assembly scene can be extracted from the segmentation result. 

\subsubsection*{Target point cloud projection} In the second function $PROJ3D()$, we project a target point cloud of base object $P_{t}$ using $I_{SEG}$ and depth image $I_{D}$ of the scene.
    
\subsubsection*{Source point cloud projection} A source point cloud $P_{s}$ with enough correspondences and an initial transformation closer to the $P_{t}$ is an essential for successful point cloud registration. To generate a $P_{s}$ with above qualities, we transform the relevant $M_{s}$ to the center of $P_{t}$ in $TFORM()$ and implement a ray casting function $RCAST()$ in simulation. The $RCAST()$ function simulates a pinhole camera looking at center of $M_{s}$ with the intrinsic and extrinsic parameters of the camera used to obtain $I_{RGB}$ and $I_{D}$ in real-time. This function generates a $P_{s}$ that has a similar partial view of the target and sufficient correspondences to $P_{t}$. This procedure eliminates the requirement to have a separate database of point clouds for registration compared to other related work \cite{SegICP,Segnet}. we record the initial transformation of $P_{s}$ as $T_w^s$.
    
\subsubsection*{Point cloud registration} In the fourth step, we implement a point cloud registration pipeline to align $P_{s}$ against $P_{t}$. 
        We estimate an initial transformation using a global registration based on Random Sample Consensus (RANSAC). RANSAC detects corresponding points in $P_{s}$ and $P_{t}$ using a nearest neighbour algorithm with Fast Point Feature Histograms (FPFH) \cite{Rusu2010} descriptors as geometrical feature inputs. Furthermore, the algorithm runs for 100,000 iterations while pruning correspondences based on edge length $E$ between any pair of corresponding points in $P_{s}$ and $P_{t}$:
         \[ || E_{s}|| > 0.09 || E_{t}||,  \] 
         \[ || E_{t}|| > 0.09 || E_{s}||,  \]
         and a point cloud distance threshold of 0.036. These values were figured out experimentally for each assembly.    
        
        A local registration based on point-to-plane ICP \cite{Point-Plane-ICP} further refines the global registration result. The refinement is based on the convergence of an objective function $L(T)$:
        \begin{equation}
        L(T) = \sum_{p \in P_{s} , q \in P_{t} } ((p - Tq) \cdot(n_{p}))^{2}, 
        \end{equation}
        where $p$ and $q$ are points in $P_{s}$ and $P_{t}$ respectively, $n_{p}$ is an estimate of normal of point $p$. The function converges when the point clouds are aligned. The resulting 6D transformation from  $P_{s}$ to $P_{t}$ in our context is equal to $T_s^b$ the estimated  6D transformation of base object relative to $P_{s}$.

\subsubsection*{Local Pose transformation}  Knowing the transformation of $P_{s}$ as $T_w^s$, the estimated transformation of base object with respect to world frame $T_w^b$ can be derived. In the final step, local assembly poses $T_b^a$ can be transferred to estimate the 6D assembly pose of each object with respect to robot base $T_w^a$ , as the output of algorithm:
    
    \begin{equation}
    T_w^b = T_s^b T_w^s,
    \end{equation}
    \begin{equation}
    T_w^a = T_b^a T_w^b.
    \end{equation}
    

\section{Evaluation}\label{sec:evaluation}

\subsection{Implementation}
 It is common to demonstrate the performance of pose estimation algorithms using existing standard datasets. However, there are no available standard datasets specifically designed for the purpose of assembly pose estimation.We therefore evaluate our method on two different object assemblies using the 3D mesh files of the assembly objects obtained from Thingiverse\footnote{\url{https://www.thingiverse.com/thing:3936460}},\footnote{\url{https://www.thingiverse.com/thing:8460}}. We generate two synthetic datasets in a format specific for assembly pose estimation in gazebo simulation environment. Table \ref{tab:fidget_table} and \ref{tab:gearbox_table} describe the helical and planetary gear assembly sets, with three assembly steps four assembly steps, respectively For ease of integration the proposed format is designed as an extension to the existing BOP format \cite{BOP} for object detection and 6D pose estimation. The data generation algorithm is presented in Alg. \ref{alg:data_gen} and the source code and links to download the datasets can be found in the git repository: \textit{\url{https://github.com/KulunuOS/6DAPose}}.
 
\begin{algorithm}[!t]
\SetAlgoLined
\SetKwInOut{KwFuncs}{Functions}
\SetKwInOut{KwInput}{Input}
\SetKwInOut{KwOutput}{Output}
\SetKwInOut{KwParams}{Parameters}
\KwParams{\\ 
        $\phi$: yaw angle of the camera\\
        $\theta$: pitch angle of the camera\\
        $s$: scale of the camera\\
}
\KwInput{CAD models of object assembly} 
\KwOutput{$I_{RGB}, I_{D}$; color and depth images \\
          $I_{SEG}$; segmentation maps, \\
          $P_{obj}$; ground truth object poses,\\ 
          $P_{cam}$; ground truth camera pose,\\
          $K_{cam}$; ground truth camera parameters \\
         }
Define and record assembly constrains\\
\ForEach{Assembly step}{
    \ForEach{$\phi, \theta, s$}{
                Record $\{I_{RGB}, I_{D}, I_s, P_{obj}, P_{cam}, K_{cam}\}$ \\ 
    }
}
\caption{\strut Assembly dataset generation}
\label{alg:data_gen}
\end{algorithm}

For our implementation we only consider primary assembly constraints, defined by the relative transformation between the base object and the assembly object $T_a^b$. This has to be individually analyzed with human expertise for distinct assemblies. The helical gear and planetary gear assemblies are constrained by three and four sets of relative poses, respectively, corresponding to their assembly steps (see Fig. \ref{fig:fidget_bbx} and \ref{fig:gearbox_bbx}). In the process, assembling order of objects, the base object for each assembling step and object information such as mesh model diameter and corners have to be analyzed. Data is recorded for each assembly step starting from the disassembled configuration. For each assembly step, a simulated RGB-D camera captures the data by following hemisphere sampling procedure \cite{Hemisphere} parameterised by yaw angle $\phi$, pitch $\theta$ and scale $s$. For the implementation, we simulate a Intel Realsense camera model with default camera parameters and capture 431 instances for each assembly step in both object assemblies. The major difference between our dataset and standard object pose estimation datasets is the inclusion of assembly steps as individual sub datasets. 
Furthermore, its important to note that contrasting to pose estimation datasets, the ground truth assembly pose for $i^{th}$ assembly step is obtained from $i+1^{th}$ assembly step. We share tools from Open3D \cite{Open3d} library to implement point cloud processing functions in our work.

\begin{table*}[!tp]
  \centering
    \caption{Helical gear assembly dataset, with four parts and three assembly steps.}
  \begin{tabular}{r|c|c|c|c|c}
    \textbf{Mesh model} & 
    \includegraphics[width=21mm]{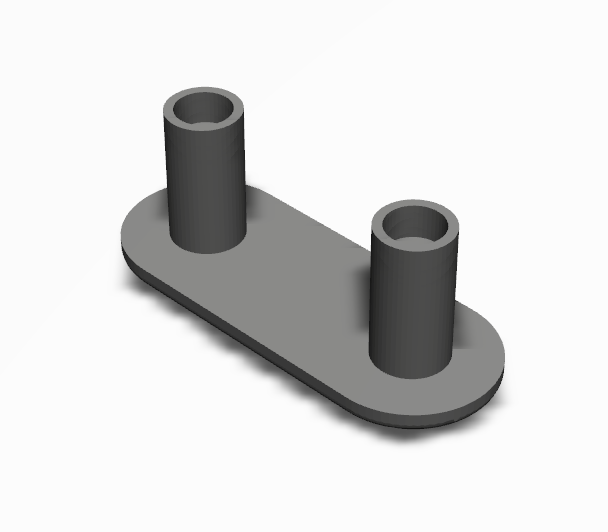} & 
    \includegraphics[width=21mm]{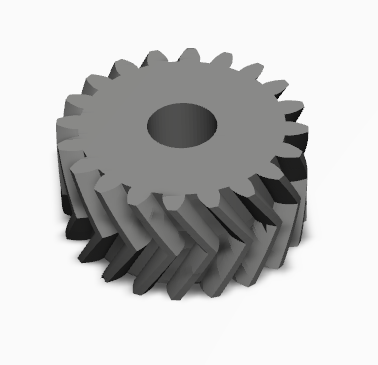} & 
    \includegraphics[width=20mm]{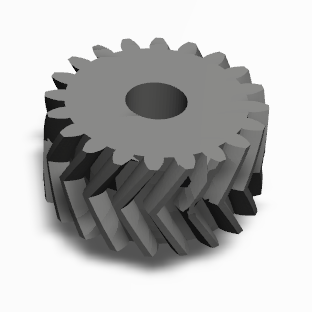} &
    \includegraphics[width=21mm]{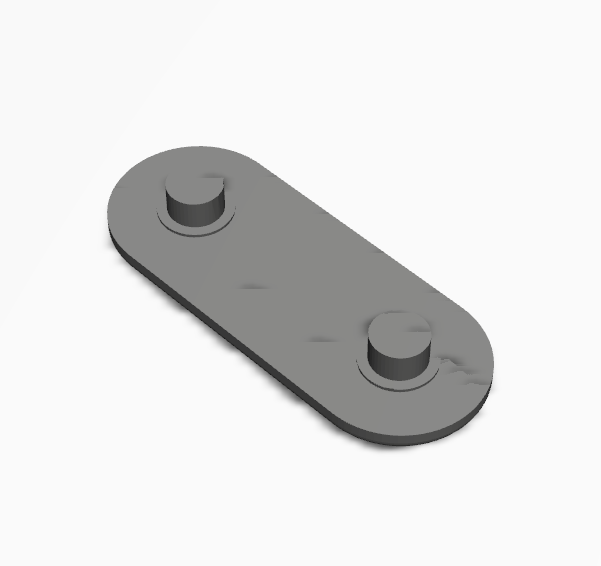} &
    \includegraphics[width=21mm]{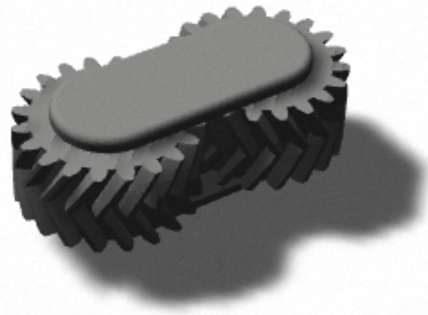} \\
    \midrule
    \textbf{Mesh name} & \text{Bottom casing} & \text{Left gear} & \text{Right gear} & \text{Top casing} & \text{Complete assembly}\\
    \textbf{Diameter [cm]} & \text{3.81} & \text{2.78} & \text{2.78} & \text{3.82} \\
  \end{tabular}

  \label{tab:fidget_table}
\end{table*}

\begin{table*}[!tp]
  \centering
    \caption{Planetary gear assembly dataset, with five major parts and four assembly steps.}
  \begin{tabular}{r|c|c|c|c|c|c}
    \textbf{Mesh model} & 
    \includegraphics[width=21mm]{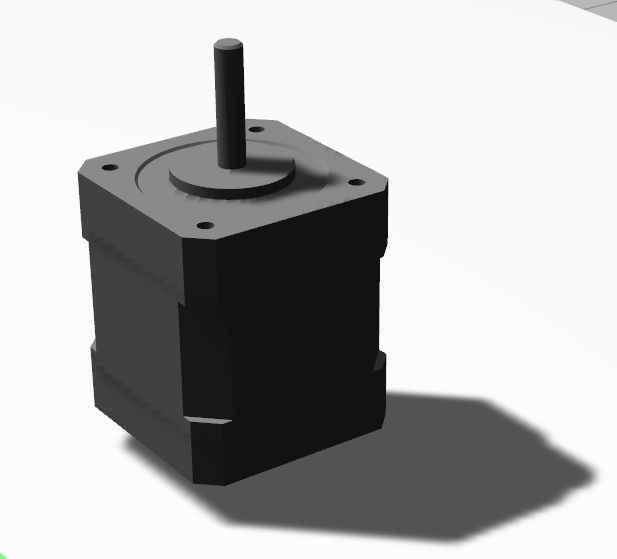} & 
    \includegraphics[width=21mm]{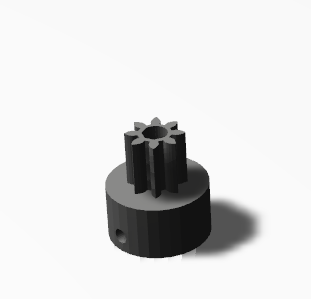} & 
    \includegraphics[width=21mm]{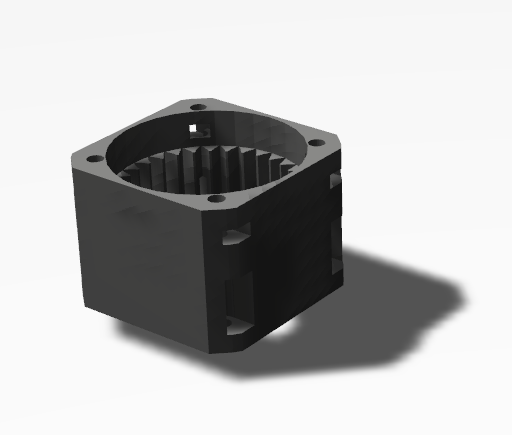} &
    \includegraphics[width=21mm]{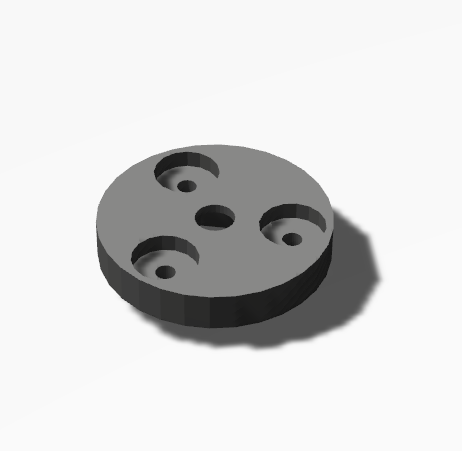} &
    \includegraphics[width=21mm]{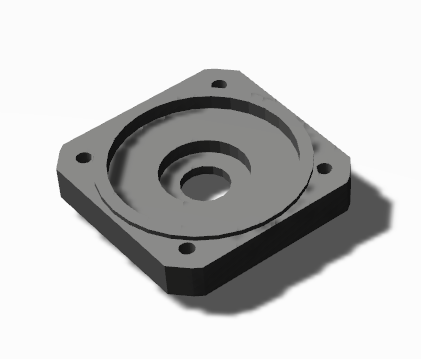} &
    \includegraphics[width=21mm]{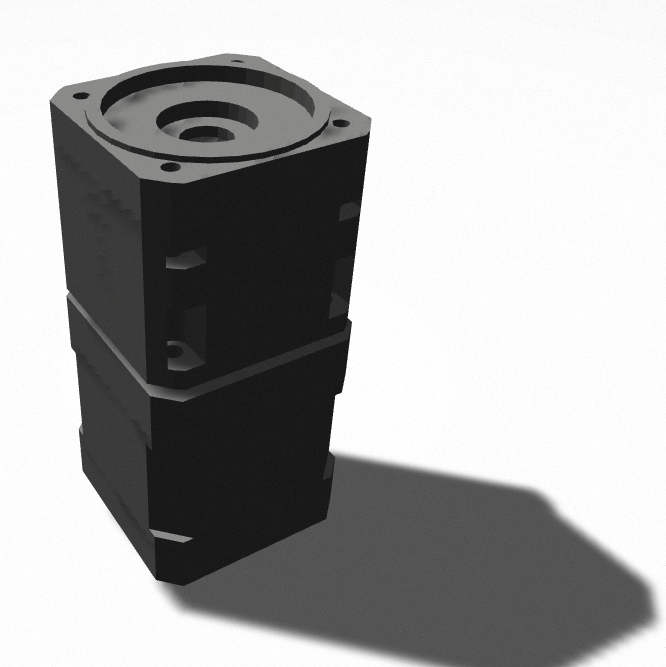} \\
    \midrule
    \textbf{Mesh name} & \text{Nema17 Motor} & \text{Sun gear} & \text{Housing} & \text{Carrier} & \text{Cover} & \text{Complete assembly}\\
    \textbf{Diameter [cm]} & \text{7.67} & \text{2.70} & \text{6.17} & \text{3.56}  & \text{5.38}\\
  \end{tabular}

  \label{tab:gearbox_table}
\end{table*}

\subsection{Metrics}
The pose estimation accuracy is directly affected by the point cloud alignment produced by the ICP registration process. Therefore for each of 431 instances in each assembly step we calculate the \textit{Fitness} and root mean square error of inliers; $l_{rmse}$ produced by point cloud registration.  \textit{Fitness} refers to the ratio between number of total inliers $l$ and total points $N$ in $P_{t}$. $l_{rmse}$ is a function that calculates the error between inliers in source  $l_{s}$ and target $l_{t}$. For an ideal point cloud alignment \textit{Fitness} should be closer to 1 and $l_{rmse}$ must be closer to 0: 

\begin{equation}
\textit{Fitness} = \frac{l}{N},
\end{equation}

\begin{equation}
l_{rmse} = \sqrt{\frac{1}{l} \sum_{j=1}^{l} \left\| l_{\text{s}_j} - l_{\text{t}_j} \right\|^2}.
\end{equation}

The symmetric nature of objects deprecates rotation and translation error as an evaluation metric for pose accuracy. Hence, we use Maximum Symmetry-aware Surface Distance \textit{MSSD} \cite{BOP} and Average Distance of model points for objects with Indistinguishable views \textit{ADI} \cite{ADI} as pose error functions to evaluate the accuracy: 

\begin{equation}
MSSD = \min_{y \in Y} \max_{x \in X} \left\| Tx - \hat{T}yx \right\|_2,
\end{equation}

\begin{equation}
ADI = \frac{1}{K} \sum_{k=1}^{K} \min_{m=1}^{M} \left\| U_k - V_m \right\|.
\label{ADI}
\end{equation}

\textit{ADI} calculates average of distances to the nearest neighbors from vertices in the ground-truth pose to vertices in the estimated pose  as described in Eqn. \ref{ADI}. $U_{k}$ and $V_{m}$ are 3D vectors describing the respective vertices where $K$ and $M$ are total number of sampled vertices. In contrast \textit{MSSD} is less dependent on sampling vertices and calculates maximum distance between surfaces as in Eqn. \ref{ADI} where $Y$ is a set of global symmetry transformations, $X$ is a set of mesh vertices, $T$ and $\hat{T}$ are ground truth and estimated 6D poses respectively. Hence \textit{MSSD} is more suitable metric for robotic manipulation tasks. The errors are calculated individually for each assembly step and does not represent a cumulative error propagation. The mean values and standard deviation for each metric is calculated for each assembly step in the whole dataset. The calculations were run in a standard computing machine with an AMD Ryzen 7 4800h CPU with cores without multi-threading. The mean time per iteration is calculated in seconds. The evaluation results of the method for the two assembly datasets are summarized in the Tables \ref{tab:figet_eval} and \ref{tab:gearbox_eval}. 

\subsection{Results}
\begin{figure*}[h]
    \centering
    \begin{subfigure}{0.27\textwidth}
        \includegraphics[width=\textwidth]{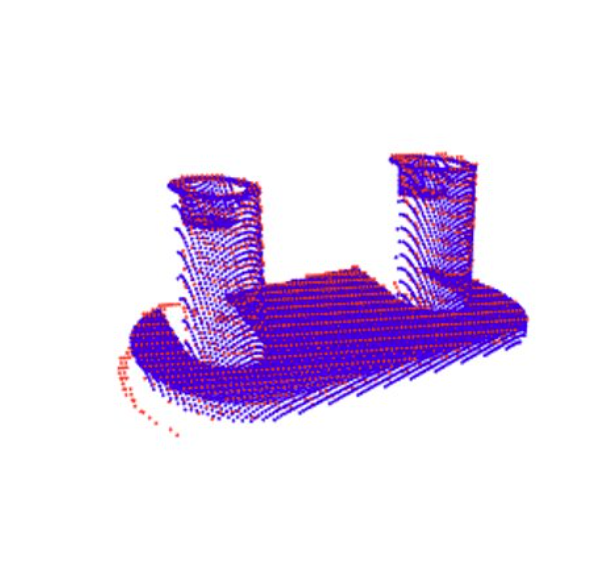}
        \caption{Base object 1}
    \end{subfigure}%
    \begin{subfigure}{0.27\textwidth}
        \includegraphics[width=\textwidth]{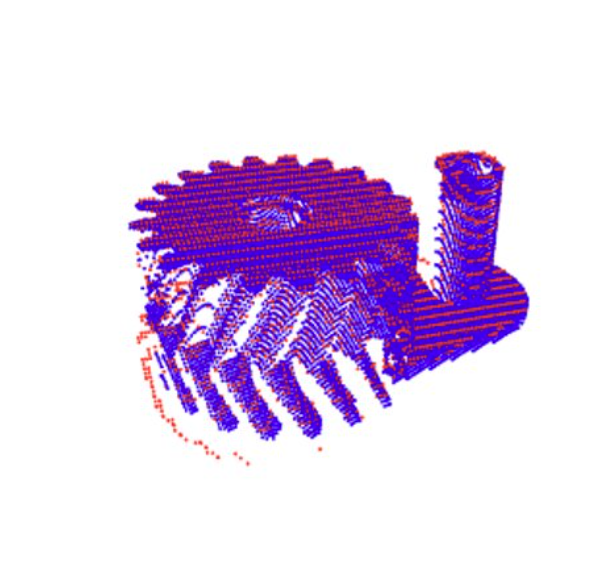}
        \caption{Base object 2}
    \end{subfigure}
    \begin{subfigure}{0.27\textwidth}
        \includegraphics[width=\textwidth]{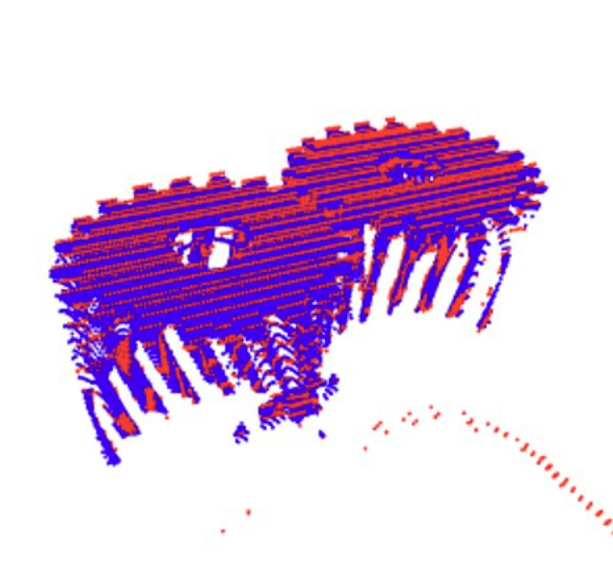}
        \caption{Base object 3}
    \end{subfigure}%
    \caption{Point cloud registration of base objects for each assembly step: bottom casing (a), bottom casing and left gear (b), bottom casing, left and right gear (c). Target (red) and source (blue) point clouds are obtained from camera and CAD, respectively.}
    \label{fig:fidget_registration}
\end{figure*}

\begin{figure*}[h]
    \centering
    \subcaptionbox{Assembly step 1 \label{fig:step1_helical}}{%
        \includegraphics[width=0.25\textwidth]{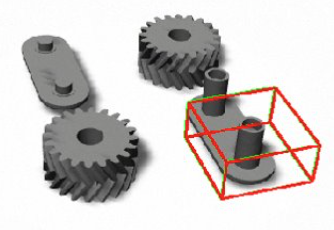}%
    }
    \subcaptionbox{Assembly step 2\label{fig:step2_helical}}{%
        \includegraphics[width=0.25\textwidth]{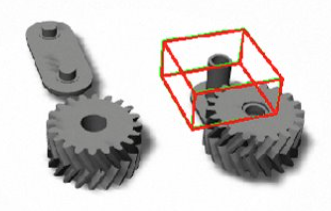}%
    }
    \subcaptionbox{Assembly step 3\label{fig:step3_helical}}{%
        \includegraphics[width=0.25\textwidth]{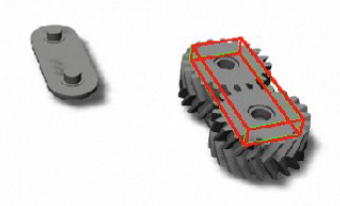}%
    }
    \caption{Estimated (green bounding box) and ground truth (red bounding box) 6D assembly poses for assembly steps 1-3 in helical gear assembly: left gear to bottom casing (a), right gear to bottom casing (b), top casing to complete assembly (c). 
    }
    \label{fig:fidget_bbx}
\end{figure*}

\begin{figure*}[h]
    \centering
    \begin{subfigure}{0.24\textwidth}
        \includegraphics[width=\textwidth]{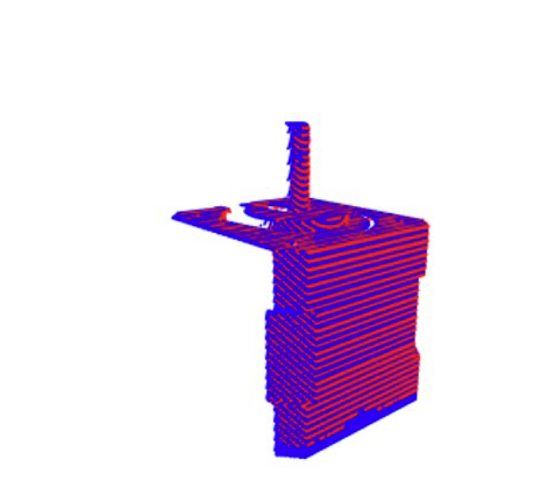}
        \caption{Base object 1}
    \end{subfigure}%
    \begin{subfigure}{0.24\textwidth}
        \includegraphics[width=\textwidth]{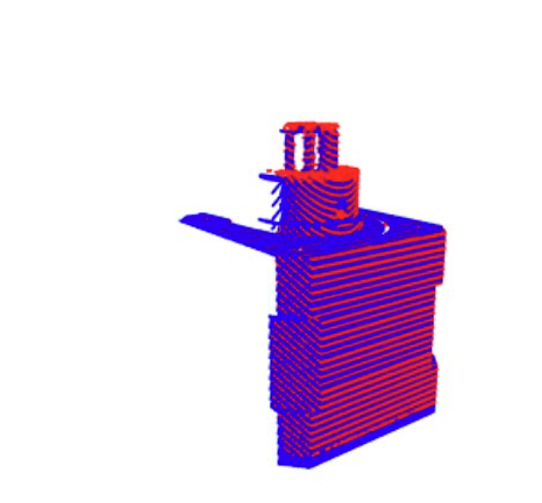}
        \caption{Base object 2}
    \end{subfigure}
    \begin{subfigure}{0.24\textwidth}
        \includegraphics[width=\textwidth]{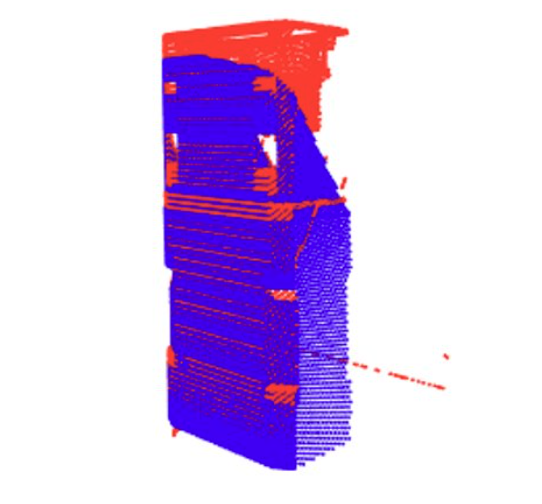}
        \caption{Base object 3}
    \end{subfigure}%
        \begin{subfigure}{0.24\textwidth}
        \includegraphics[width=\textwidth]{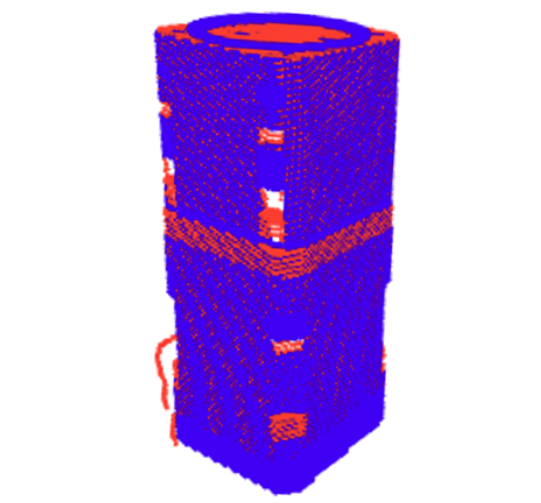}
        \caption{Base object 4}
    \end{subfigure}%
    \caption{Point cloud registration of base objects for each assembly step: motor (a), motor and sun gear (b), motor, sun gear and housing (c),  motor, sun gear, housing and cover (d). Target (red) and source (blue) point clouds are obtained from camera and CAD, respectively.
    }
    \label{fig:gearbox_registration}
\end{figure*}

\begin{figure*}[h]
    \centering
    \subcaptionbox{Assembly step 1\label{fig:step1_planetary}}{%
        \includegraphics[width=0.24\textwidth]{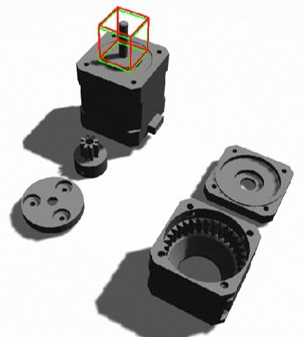}%
    }
    \subcaptionbox{Assembly step 2\label{fig:step2_planetary}}{%
        \includegraphics[width=0.24\textwidth]{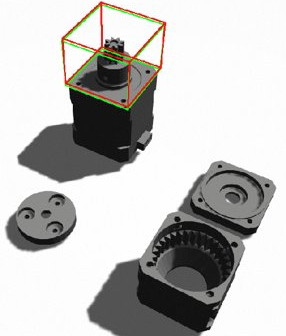}%
    }
    \subcaptionbox{Assembly step 3\label{fig:step3_planetary}}{%
        \includegraphics[width=0.24\textwidth]{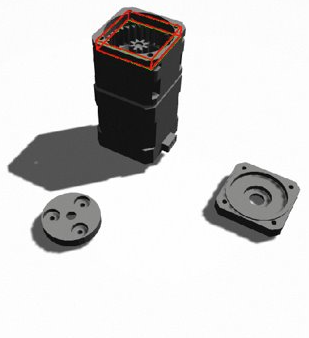}%
    }
    \subcaptionbox{Assembly step 4\label{fig:step4_planetary}}{%
        \includegraphics[width=0.24\textwidth]{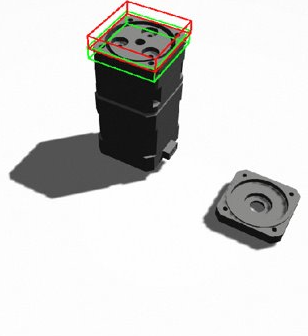}%
    }

    \caption{Estimated (green bounding box) and ground truth (red bounding box) 6D assembly poses for assembly steps 1-4 in planetary gear assembly: sun gear to motor (a), housing to motor (b), carrier to housing (c), cover to complete assembly (d).
    }
    \label{fig:gearbox_bbx}
\end{figure*}

We present evaluation results for two simulated gear assembly datasets illustrated in Tables \ref{tab:figet_eval} and \ref{tab:gearbox_eval}. The step-wise propagation of point cloud registration of base objects between target (red) and source (blue) point clouds are illustrated in Fig. \ref{fig:fidget_registration} and \ref{fig:gearbox_registration}, demonstrating well-matching overlap in all cases. The estimated (green) and ground truth (red) 6D assembly poses for each assembly steps are displayed using bounding box representation in Fig. \ref{fig:fidget_bbx} and \ref{fig:gearbox_bbx}. A statistical analysis of all estimated assembly 6D poses are summarized in Tables \ref{tab:figet_eval} and \ref{tab:gearbox_eval}. The analysis shows that when it is possible to achieve good \textit{Fitness} values closer to 1  with $l_{rmse}$ closer to 0, the assembly poses can be estimated with a high accuracy according to the pose metrics. Values closer to 0 in \textit{MSSD} explains that surfaces align well when an assembly mesh is rendered for an estimated pose, as compared to the ground truth pose. Similarly in \textit{ADI}, values closer to 0 suggests that distances between vertices of an assembly mesh in estimated pose are closer to that of the ground truth result.

One exception is the assembly step 4 for the planetary gear dataset, which has a comparatively bigger pose error even with consistent \textit{Fitness} values. The reason for this is the occlusions forced on assembly step 4, which is located inside assembly object 3. An absence of important assembly surfaces introduces a pose estimation error even when base object point clouds fit well. 

A second observation is that the time for pose estimation increases with the increase of the assembly steps. This is due to an increase in the number of points in the source point cloud, hence increasing the computational load of the point cloud registration approach. In our cases, we sample 30,000 points for each assembly object in the dataset.



\begin{table*}[h!]
\centering
\caption{Evaluation metrics for 6D assembly pose estimation on the helical gear dataset}
\begin{tabular}{c|ccccccccc}
\toprule
Assembly step        &  \multicolumn{2}{c}{\textit{Fitness}}  &  \multicolumn{2}{c}{$l_{rmse}$}   & \multicolumn{2}{c}{\textit{ADI}} &  \multicolumn{2}{c}{\textit{MSSD}} &  time [sec] \\
  \midrule
     &       mean             &  stdv    &      mean          &   stdv    & mean     & stdv         &  mean      & stdv         & mean\\
  1  &      0.995184 &      0.062205 &      0.000429 &      0.000097 &  0.000604 &  0.004411 &   0.001075 &   0.005385 &   0.505338 \\
  2  &      0.999543 &      0.000571 &      0.000436 &      0.000073 &  0.000374 &  0.000036 &   0.000741 &   0.000156 &   0.562895 \\
  3  &      0.999333 &      0.000693 &      0.000444 &      0.000085 &  0.000510 &  0.000057 &   0.000778 &   0.000425 &   0.646653 \\
\bottomrule
\end{tabular}

\label{tab:figet_eval}
\end{table*}

\begin{figure*}[h!]
\sbox\twosubbox{%
  \resizebox{\dimexpr.95\textwidth}{!}{%
    \includegraphics[height=2.0cm]{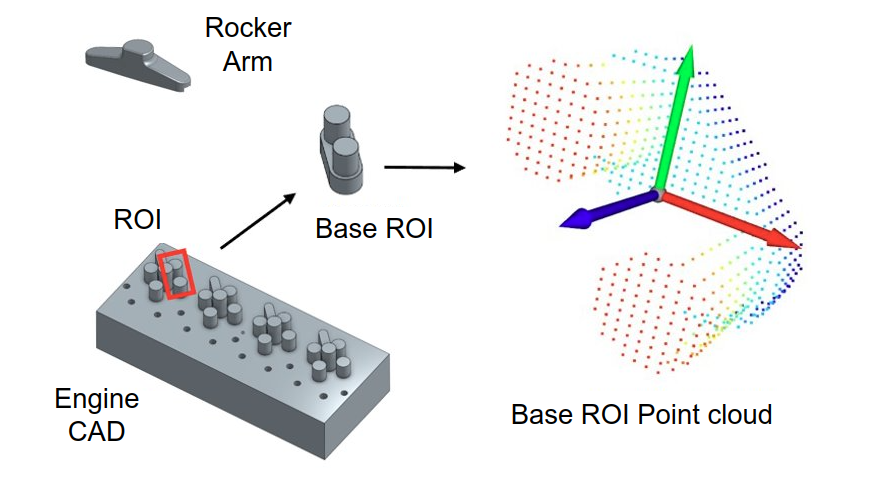}%
    \includegraphics[height=2.0cm]{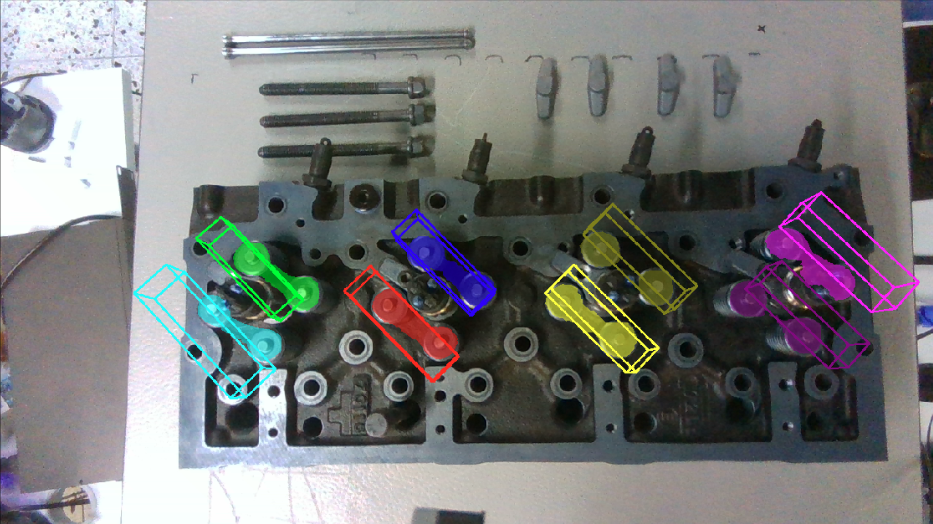}%
  }%
}
\setlength{\twosubht}{\ht\twosubbox}
\centering
\subcaptionbox{\label{fig:diesel_CAD}}{%
  \includegraphics[height=\twosubht]{figures/engine_results/CAD_explanation.png}%
}\quad
\subcaptionbox{\label{fig:diesel_bbx}}{%
  \includegraphics[height=\twosubht]{figures/engine_results/engine_result_bbx.png}%
}\quad
\caption{Diesel engine assembly pose estimation results for eight rocker arms. Point cloud registration utilizes base regions of interest (ROI) on the assembly surface (a), obtained from semantic segmentation (b). The eight estimated assembly poses are highlighted with 3D bounding boxes (b).\label{fig:diesel_result}}
\end{figure*}

To observe the sim-to-real gap, we also evaluated our method on an industrial use case, for the estimation of eight assembly poses of rocker arms on a Diesel engine (see Fig. \ref{fig:diesel_result}). 
The base object for this application is a region of interest (ROI) on the assembly surface. Henceforth we use the term base ROI to describe it. A semantic segmentation module was trained on Detectron2 \cite{detectron2} Mask-RCNN architecture to detect base ROIs as shown in Fig. \ref{fig:diesel_bbx}. The module was trained on 40 captured images and manually labelled annotations. The estimated base ROI and bounding box for estimated pose are illustrated in similar colors. The results emphasize that the method can produce feasible assembly poses for different regions of the surface while having less consistency compared to simulated data due to occlusions from the engine surface. Unlike in a simulated environment, ground truth information is unavailable for real life applications. Therefore, the accuracy of the estimated assembly pose has to be visually inspected. Due to absence of precise CAD models of the engine assembly we use an approximated CAD model to generate base ROI as explained in Fig. \ref{fig:diesel_CAD}. This adversely affects the point cloud registration by reducing point correspondences. However, the method generates feasible assembly poses when there are sufficient correspondences. The number of correspondences can be improved by changing the angle of the camera to capture as many points on the ROI as possible. 
This can be enabled by changing the viewing angle of the camera as placed on the end-effector of the robot. As visible in Fig. \ref{fig:diesel_bbx}, the bounding boxes align well on ROI surfaces with more correspondences (red, blue, yellow) and weaker alignments on incomplete ROI segments with less correspondences (cyan, purple). The observations suggest that surface regions located at the center and corners of the field of view tend to capture less ROI surfaces compared to other regions. The estimated poses are sufficient for assembling, considering the magnetic force between rocker arm and surface in the application.




\begin{table*}[h!]
\centering
\caption{Evaluation metrics for 6D assembly pose estimation on the planetary gear dataset}
\begin{tabular}{c|ccccccccc}
\toprule
Assembly step        &  \multicolumn{2}{c}{\textit{Fitness}}  &  \multicolumn{2}{c}{$l_{rmse}$}   & \multicolumn{2}{c}{\textit{ADI}} &  \multicolumn{2}{c}{\textit{MSSD}} &  time [sec] \\
\midrule
               &      mean             &  stdv    &      mean          &   stdv    & mean     & stdv         &  mean      & stdv         & mean\\
  1            &      0.999880 &      0.000417 &      0.000548 &      0.000109 &  0.000528 &  0.000045 &   0.001425 &   0.000087 &   1.329737 \\
  2            &      0.990592 &      0.096101 &      0.000618 &      0.000122 &  0.002384 &  0.019576 &   0.003604 &   0.023681 &   1.359196 \\
  3            &      0.999851 &      0.000489 &      0.000534 &      0.000146 &  0.000427 &  0.000619 &   0.000796 &   0.000964 &   2.369656 \\
  4            &      0.997401 &      0.048216 &      0.001335 &      0.000199 &  0.003576 &  0.008271 &   0.006678 &   0.009545 &   2.396376 \\
\bottomrule
\end{tabular}

\label{tab:gearbox_eval}
\end{table*}

\section{Discussion}\label{sec:discussion}
The proposed framework is capable of estimating a 6D assembly pose for an object assembly without initializing source point clouds at a predefined constraint, unlike Zeng et al. \cite{amazon} or other work that require creating a model library \cite{SegICP}. Furthermore, calculating two separate metrics for point cloud registration step and pose estimation provides a feedback on the quality of captured depth information, pose initialization and point cloud registration parameters. Unlike work that only estimates object poses \cite{xu2019,wang2019pose}, which is more suited for bin picking and sorting tasks, the proposed framework can put together an object assembly while maintaining quality control. 

Although the proposed method estimates assembly poses accurately for object assemblies with few objects, the computational load increases for complex assemblies with multiple objects, due to large number of points accumulated in the registration step. Similarly, the absence of precise CAD models decreases the number of correspondences in derived source point clouds which adversely affects point cloud registration. Furthermore, as a purely geometrical approach, the method is not robust to occlusions and obstructions in the scene. Occlusions and obstructions eliminate points, which represent important features and affect correct point cloud alignments. An example is when assembling or placing an object internally on a place surrounded by walls or surfaces such as plumbing, box packaging etc. In such cases, although a point cloud registration achieves a good fitness score, the resulting assembly pose may be less accurate.

As a future improvement of this work it would be appropriate to have a learning-based pose estimation module before the point cloud registration step. A key point-based approach would reduce the computational load by ruling out the necessity to deal with a complete point cloud with large number of points. It will make it possible to use the framework on complex object assemblies with a large number of objects without compromising the inference time. Estimated assembly poses can also be used to prune unfeasible object grasp candidates in grasp detection networks. Moreover, complex robot manipulation tasks that involve additional constraints will be considered, as the next step of assembly pose estimation. This includes specific placement direction or motions, as found in insertion tasks, such as peg-in-hole or clamping.

\section{Conclusion}\label{sec:conclusion}
This work proposed an assembly pose estimation method for robot manipulation and assembly tasks, utilizing semantic segmentation and point cloud registration between target and source point clouds. The work also presented a synthetic data generation pipeline as an improvement to existing object pose estimation datasets. Our approach is evaluated with suitable metrics on two simulated gear assembly datasets, which indicates that point cloud registration is well capable of estimating 6D assembly poses for object assemblies. In addition, we demonstrate the approach on a real industrial assembly task, i.e., Diesel engine assembly, which verifies that feasible assembly poses can be estimated for real applications. 

\section*{Acknowledgements}

Project funding was received from Helsinki Institute of Physics' Technology Programme (project; ROBOT) and European Union's Horizon 2020 research and innovation programme, grant agreement no. 871252 (METRICS). 

\bibliographystyle{IEEEtran}
\bibliography{IEEEabrv,refs}

\end{document}